\documentclass{ieeeaccess}
\usepackage{cite}
\usepackage{amsmath,amssymb,amsfonts}
\usepackage{algorithmic}
\usepackage{graphicx}
\usepackage{textcomp}
\usepackage{booktabs}
\usepackage{bm}
\makeatletter
\AtBeginDocument{\DeclareMathVersion{bold}
\SetSymbolFont{operators}{bold}{T1}{times}{b}{n}
\SetSymbolFont{NewLetters}{bold}{T1}{times}{b}{it}
\SetMathAlphabet{\mathrm}{bold}{T1}{times}{b}{n}
\SetMathAlphabet{\mathit}{bold}{T1}{times}{b}{it}
\SetMathAlphabet{\mathbf}{bold}{T1}{times}{b}{n}
\SetMathAlphabet{\mathtt}{bold}{OT1}{pcr}{b}{n}
\SetSymbolFont{symbols}{bold}{OMS}{cmsy}{b}{n}
\renewcommand\boldmath{\@nomath\boldmath\mathversion{bold}}}
\makeatother

\def\BibTeX{{\rm B\kern-.05em{\sc i\kern-.025em b}\kern-.08em
    T\kern-.1667em\lower.7ex\hbox{E}\kern-.125emX}}

\begin{document}
\history{}
\doi{}

\title{ICAS: IP-Adapter and ControlNet-based Attention Structure for Multi-Subject Style Transfer Optimization}
\author{\uppercase{Fuwei Liu}\authorrefmark{1}}

\address[1]{Northeastern University at Qinhuangdao, CO 066004 CN}
\markboth
{F. Liu: ICAS: IP-Adapter and ControlNet-based Attention Structure for Multi-Subject Style Transfer Optimization}
{F. Liu: ICAS: IP-Adapter and ControlNet-based Attention Structure for Multi-Subject Style Transfer Optimization}

\corresp{Corresponding author: Fuwei Liu (e-mail: 202412277@stu.neuq.edu.cn).}

\begin{abstract}
Generating multi-subject stylized images remains a significant challenge due to the ambiguity in defining style attributes (e.g., color, texture, atmosphere, and structure) and the difficulty in consistently applying them across multiple subjects. Although recent diffusion-based text-to-image models have achieved remarkable progress, existing methods typically rely on computationally expensive inversion procedures or large-scale stylized datasets. Moreover, these methods often struggle with maintaining multi-subject semantic fidelity and are limited by high inference costs.
To address these limitations, we propose ICAS (IP-Adapter and ControlNet-based Attention Structure), a novel framework for efficient and controllable multi-subject style transfer. Instead of full-model tuning, ICAS adaptively fine-tunes only the content injection branch of a pre-trained diffusion model, thereby preserving identity-specific semantics while enhancing style controllability. By combining IP-Adapter for adaptive style injection with ControlNet for structural conditioning, our framework ensures faithful global layout preservation alongside accurate local style synthesis.
Furthermore, ICAS introduces a cyclic multi-subject content embedding mechanism, which enables effective style transfer under limited-data settings without the need for extensive stylized corpora. Extensive experiments show that ICAS achieves superior performance in structure preservation, style consistency, and inference efficiency, establishing a new paradigm for multi-subject style transfer in real-world applications.
\end{abstract}


\titlepgskip=-21pt

\maketitle

\section{Introduction} \label{sec:introduction}

Diffusion-based text-to-image generation models have demonstrated remarkable capabilities in synthesizing high-quality images conditioned on various inputs. Among them, image style transfer—aiming to combine the semantic content of a source image with the artistic or structural style of a reference—remains a long-standing and challenging task. The core difficulty lies in preserving multi-subject semantics while achieving consistent style adaptation across diverse elements such as color, texture, layout, and scene attributes~\cite{jing2019neural, deng2020arbitrary}.

Recent advances have explored diffusion models for human-centric generation, including pose-guided generation~\cite{shen2024imagpose}, virtual dressing~\cite{shen2024imagdressing}, and talking face synthesis~\cite{shen2025long}, leveraging the structural controllability and high-fidelity rendering of pretrained diffusion backbones. However, most of these works focus on identity or pose control, while multi-subject style transfer, especially under limited data settings, remains underexplored.

Existing methods fall into two broad categories. The first line of work avoids training and relies on image inversion or feature injection from pre-trained models. Approaches like DDIM inversion~\cite{song2020denoising}, Plug-and-Play~\cite{tumanyan2023plug}, and StyleID~\cite{chung2024style} perform content/style encoding via inversion pipelines, injecting them into the denoising process using attention or projection modules. However, these methods are computationally intensive, prone to identity drift, and often fail under multi-subject scenarios due to inversion inaccuracies~\cite{mokady2023null}. IP-Adapter-based approaches, such as InstantStyle~\cite{wang2024instantstyle}, improve runtime efficiency but struggle with maintaining subject identity when multiple entities are present.
The second category depends on training models using large-scale stylized datasets. Methods like ZipLoRA~\cite{shah2024ziplora}, B-LoRA~\cite{frenkel2024implicit}, and StyleShot~\cite{gao2024styleshot} disentangle style and content representations via dedicated training. While they improve control, they require expensive fine-tuning and dataset curation, limiting practicality in open-world or data-scarce scenarios.

To address these challenges, we propose ICAS (IP-Adapter and ControlNet-based Attention Structure), a novel framework for controllable multi-subject style transfer. Rather than retraining the entire model or performing expensive inversions, ICAS selectively fine-tunes the content injection branch of a pre-trained diffusion model, effectively preserving subject semantics while enabling style adaptation. Inspired by recent advances in modular control and multi-condition generation~\cite{shen2024boosting, shen2023advancing}, we integrate IP-Adapter for flexible style injection and ControlNet for spatial structure conditioning. This hybrid mechanism ensures both global layout preservation and local style fidelity without reliance on large stylized datasets.
Extensive experiments validate that ICAS achieves superior performance in structure preservation, style consistency, and inference efficiency compared to both inversion-based and data-intensive baselines. Our framework establishes a new paradigm for lightweight and controllable multi-subject style transfer, particularly suited for real-world applications with limited annotations.

Our main contributions are as follows:
\begin{itemize} 
\item We propose a lightweight fine-tuning strategy that selectively updates the content cross-attention branch, preserving semantic details across multiple subjects while maintaining model efficiency. 
\item We introduce a novel integration of IP-Adapter and ControlNet to jointly enable style fidelity and structural controllability, achieving robust style transfer without extensive stylized data. 
\item We demonstrate the effectiveness of our method across diverse multi-subject scenarios, achieving state-of-the-art performance in terms of structure preservation, style consistency, and inference speed. \end{itemize}

\section{Related Work}\label{sec:rw}
\subsection{Text-to-Image Diffusion Models}
In recent years, the field of text-to-image conversion has continued to show strong vitality. Among them, the diffusion model has attracted widespread attention in the field of text-to-image conversion due to its powerful generation ability, as can be seen in the early work of Dhariwal \& Nichol (2021) \cite{dhariwal2021diffusion} and Ramesh et al. (2022) \cite{ramesh2022hierarchical}. GLIDE \cite{nichol2021glide} introduced the cascaded diffusion model architecture. Building on this, DALL-E 2 \cite{ramesh2022hierarchical} leveraged the diffusion model to condition image embeddings and trained the previous model to generate image embeddings from text prompts. This breakthrough enabled not only text-to-image generation, but also image-to-image translation tasks. Imagen further advanced this technology by using T5 \cite{raffel2020exploring}, a large Transformer model pre-trained on pure text data, as the text encoder of the diffusion model, greatly improving text understanding capabilities. Re-Imagen \cite{chen2022re} subsequently improved on this by introducing an information retrieval module that greatly improved the fidelity of generated images for rare or unseen entities.

Stable diffusion (SD) \cite{rombach2022high} represents another step forward, based on the latent diffusion model that operates directly in latent space instead of pixel space. This innovation enables SD to generate high-resolution images more efficiently. 
Stable diffusion XL (SDXL) \cite{podell2023sdxl}, one of the updated versions after SD, adds more attention blocks and larger cross-attention contexts, and introduces a refinement model to further improve the visual fidelity of generated samples. To more precisely control the generation process, Composer \cite{huang2023composer} introduces a joint fine-tuning strategy based on a pre-trained diffusion model and image embedding conditions, which supports conditional guidance for various image generation. Popular models such as ControlNet \cite{zhang2023inversion}, T2Iadapter \cite{mou2024t2i}, and IP-Adapter \cite{ye2023ip} introduce additional image conditions to enhance controllability. These models use complex feature extraction methods and integrate these features into well-designed modules to achieve layout control.
\subsection{Image Style Transfer}  
Image style transfer (often referred to as style transfer) aims to impart an artistic style to a target image. Early approaches in style transfer, including optimization-based and inference-based methods, faced limitations related to computational efficiency and the variety of transferable styles \cite{gatys2016image}, \cite{chen2017stylebank}, \cite{dumoulin2016learned}. Adaptive Instance Normalization (AdaIN) introduced a significant advance \cite{huang2017arbitrary}, which elegantly separated content and style from deep features, inspiring numerous subsequent methods that use statistical measures of deep feature distributions \cite{chen2021artistic}, \cite{hertz2024style}. Transformer-based architectures such as StyleFormer \cite{wu2021styleformer} and StyTR2 \cite{deng2022stytr2} improved the fidelity of style transfer but remained restricted to limited types of stylization, primarily color and stroke transfer. Furthermore, approaches such as InST \cite{zhang2023inversion} and inversion-based methods such as DDIM inversion \cite{song2020denoising} have emerged, utilizing attention manipulation to control stylization. However, these techniques suffer from information loss and an increased inference time due to latent inversion processes, which restrict the retention of fine-grained details such as textures and colors \cite{hertz2024style}, \cite{qi2024deadiff}.

Recent studies have explored tuning-free lightweight adapters, such as IP-Adapter \cite{ye2023ip}, Style-Adapter \cite{wang2023styleadapter}, StyleAlign \cite{hertz2024style}, and Swapping Self-Attention \cite{jeong2024visual}, to inject style information into the diffusion process directly through self-attention or cross-attention mechanisms. However, these adapters occasionally encounter content leakage and fail to effectively disentangle style from content, limiting their performance. Alternatively, fine-tuning-based methods, such as Zip-LoRA \cite{shah2024ziplora}, B-LoRA \cite{frenkel2024implicit}, and StyleShot \cite{gao2024styleshot}, deliver improved stylization quality, but require substantial computational resources and struggle with flexibility across different style domains.

\subsection{Evolving Style Transfer with Integrated Attention Mechanisms}
To address these issues, this study proposed ICAS: an attention structure based on IP-Adapter and ControlNet, which is an optimized solution for combining IP-Adapter and ControlNet to achieve effective multi-subject style transfer. By combining the advantages of IP-Adapter and ControlNet, the attention mechanism is improved, and precise control of multi-subject content and overall style is achieved without a lot of fine-tuning or inversion, effectively improving the accuracy and stability of style transfer in multi-subject scenarios, and significantly improving the flexibility and quality of real-world style transfer tasks.

\begin{figure*}[t]
    \centering
    \includegraphics[width=1\linewidth]{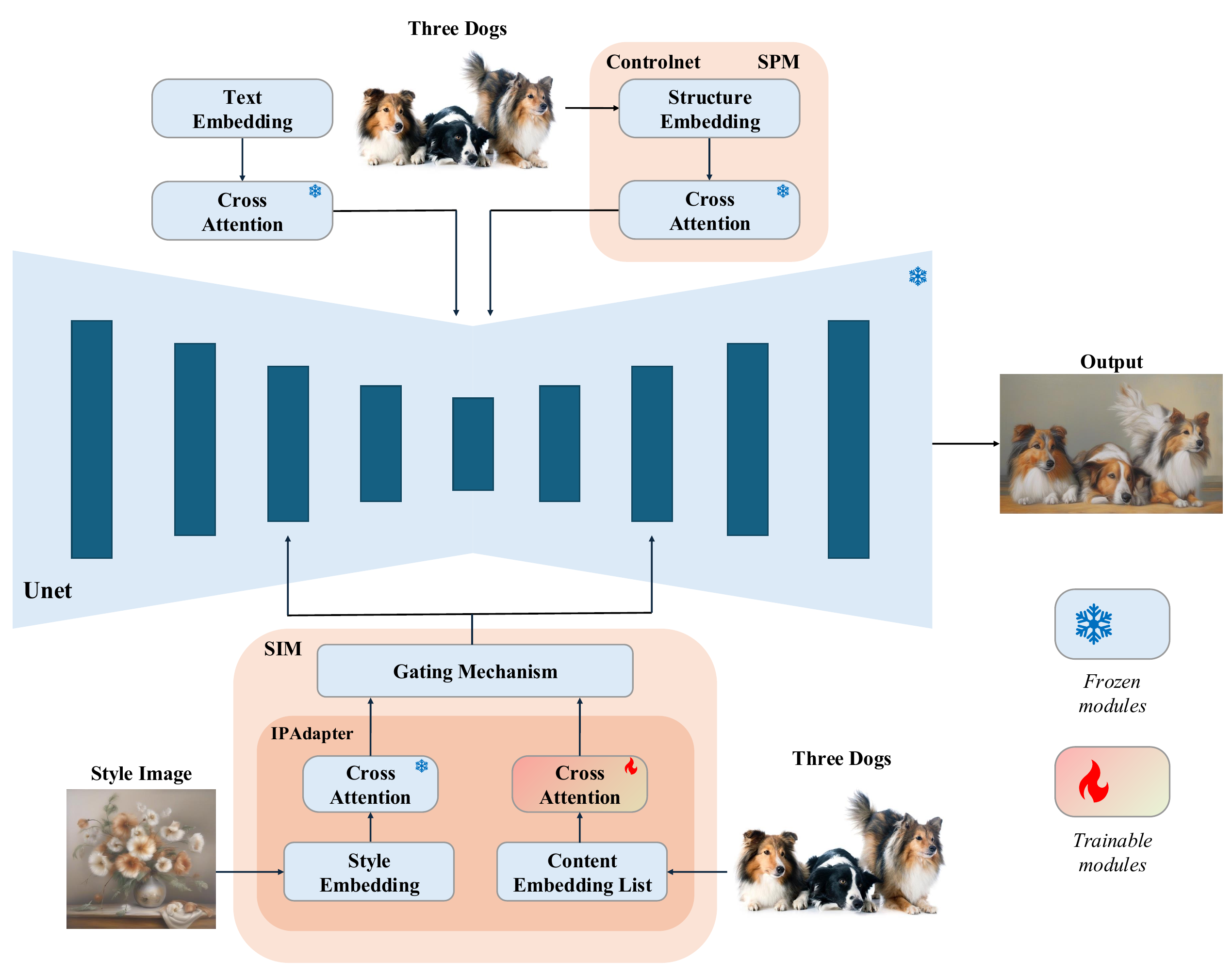}
\caption{\textbf{The overall architecture of our proposed ICAS.} Combined with pre-trained IP-Adapter (SIM) for style injection, and ControlNet-based structure preservation module (SPM).
Specifically, style image embedding and content embedding list are injected through different cross-attention paths in IP-Adapter to ensure multi-subject content feature and style feature fusion (“SIM”), while the ControlNet branch (“SPM”) receives structural conditions (e.g., edges or depth) to maintain global layout. We also fine-tune iPadAdapter on the content graph by training cross-attention on iPadAdapter path using a small dataset of our own choice. Then, the unified information flows through the diffusion U-Net to generate the final stylized image, which preserves both subject identity and structural consistency.}

\label{fig:ICAS}
\end{figure*}
\section{Proposed Method}\label{sec:method} 
\subsection{Overview}
In this section, we present \textbf{ICAS} (\textbf{I}P-Adapter and \textbf{C}ontrolNet-based \textbf{A}ttention \textbf{S}tructure), a lightweight and tuning-free method that achieves style-consistent, structure-preserving multi-subject image generation. ICAS consists of two main components: the Style Injection Module (SIM) and the Structure Preservation Module (SPM). The key idea is to decouple style and structure guidance and inject both through carefully designed adapter branches, enhancing flexibility and fidelity without fine-tuning the base diffusion model. Notably, we adopt a \emph{hybrid} strategy regarding IP-Adapter's learned blocks: the \emph{style injection path} (Section~\ref{sec:sim}) is fully inherited from a large-scale pre-trained IP-Adapter and kept frozen, while we \emph{only update a small content injection sub-block} with minimal parameters when training on limited data. This design is characterized by a reduction in computational overhead, coupled with an assurance of adaptability across multiple subjects.

Given a content image $C$ and a style reference image $R$, our objective is to synthesize an output image $\hat{I}$ that retains the structure of $C$ while adopting the visual style of $R$. Additionally, a structural condition $S$ (e.g. pose, edge) can optionally be provided.

We denote:
\begin{itemize}
    \item $\boldsymbol{e}_C = E(C)$: content image embedding from a frozen CLIP image encoder.
    \item $\boldsymbol{e}_R = E(R)$: style image embedding.
    \item $\boldsymbol{e}_S = E_S(S)$: structure condition embedding from a dedicated encoder $E_S$.
\end{itemize}

The aforementioned embeddings are integrated into the U-Net of a pre-trained diffusion model through the implementation of cross-attention and residual feature injection methodologies.

\subsection{Style Injection Module (SIM)}\label{sec:sim}  
\textbf{Motivation:} Transferring style accurately and flexibly is essential for controllable image generation. Inspired by IP-Adapter~\cite{ye2023ip}, we propose a modified style injection method that introduces a gating term to improve adaptability between content and style.

Following IP-Adapter, we use cross-attention layers to inject style features. For a U-Net query feature $\mathbf{Q} \in \mathbb{R}^{N \times d}$, we project the style embedding $\boldsymbol{e}_R \in \mathbb{R}^d$ into key and value matrices:
\begin{equation}
\mathbf{K}_R = W_K \boldsymbol{e}_R, \quad \mathbf{V}_R = W_V \boldsymbol{e}_R.
\end{equation}
where $W_K, W_V \in \mathbb{R}^{d \times d}$ are learnable matrices.

The standard attention output is:
\begin{equation}
\mathbf{A}_R = \text{softmax}\left(\frac{\mathbf{Q} \mathbf{K}_R^\top}{\sqrt{d}}\right) \mathbf{V}_R.
\end{equation}

We introduce a gating mechanism to control the contribution of style characteristics based on the similarity between $\boldsymbol{e}_C$ and $\boldsymbol{e}_R$. Let:
\begin{equation}
\boldsymbol{g} = \sigma\left(W_g (\boldsymbol{e}_C \cdot \boldsymbol{e}_R) + b_g\right).
\end{equation}
where $\cdot$ denotes element-wise multiplication, $\sigma$ is the sigmoid function, and $W_g \in \mathbb{R}^{d \times d}$, $b_g \in \mathbb{R}^d$ are learnable parameters.

The final injected feature is:
\begin{equation}
\mathbf{F}_{\text{sim}} = \alpha \cdot \mathbf{A}_R + (1 - \alpha) \cdot \mathbf{Q} + \boldsymbol{g}.
\end{equation}
where $\alpha \in [0, 1]$ is a user-controllable scalar. This formulation enables fine-grained modulation of style strength while preserving the base semantic content.

To achieve good results and avoid expensive retraining of the entire IP adapter, we \emph{freeze} the style block parameters that project $\boldsymbol{e}_R$ into $(\mathbf{K}_R, \mathbf{V}_R)$. Consequently, SIM reuses the pre-trained style injection capability, while only gating or content-related parameters get updated if needed. This partitioning aligns with the method's aim: consistent style infusion from large-scale pretrained knowledge, plus minimal overhead for multi-subject adaptation.

\textbf{Benefit:} In comparison to naive embedding fusion, our gated injection provides dynamic adaptability to image content, mitigating issues such as style leakage or over-stylization. This aligns with our objective of controllable multi-subject image generation.

\subsection{Structure Preservation Module (SPM)}  
\textbf{Motivation:} Style transfer often leads to distortion of spatial layouts, especially in multi-subject scenarios. To address this, we incorporate a structure preservation branch based on ControlNet~\cite{zhang2023inversion}, which injects condition features into intermediate U-Net layers.

Given structure input $S$, we extract a feature map $\mathbf{F}_S \in \mathbb{R}^{H \times W \times d}$ using a frozen encoder $E_S$. A small projection network $\phi$ transforms it into a residual map:
\begin{equation}
\mathbf{R}_S = \phi(\mathbf{F}_S).
\end{equation}

The residual is added to the U-Net features at selected blocks:
\begin{equation}
\mathbf{F}^{(i)}_{\text{unet}} \leftarrow \mathbf{F}^{(i)}_{\text{unet}} + \gamma \cdot \mathbf{R}_S.
\end{equation}
where $\gamma$ is a scaling parameter controlling the influence of structure.

\textbf{Benefit:} This residual injection guarantees that the generated image remains aligned with the spatial layout of the content, a crucial consideration for scenes with multiple subjects, interactions, or structural details.

\section{Experiment and Analysis}\label{sec:exp}  

In this section, we present a comprehensive evaluation of our proposed ICAS method (IP-Adapter and ControlNet-based Attention Structure) for multi-subject style transfer. The present study will commence with the delineation of the experimental methodology and baseline approaches. This will be followed by detailed qualitative and quantitative comparisons. Furthermore, we execute a series of ablation studies, which are meticulously designed to systematically analyze the contribution of each component within our framework. All experiments were conducted on an NVIDIA Tesla V100 GPU with 32GB of memory.

\subsection{Implementation Details}
\noindent\textbf{Implementation on SDXL.}
The integration of our ICAS approach into Stable Diffusion XL (SDXL) has been demonstrated to enhance image generation. In particular, the IP-Adapter (SDXL 1.0) mechanism was adopted for style injection. The present study explored two settings:

\begin{enumerate}
    \item Using a pre-trained IP-Adapter: We directly employed the publicly available weights, muting out all non-style blocks (e.g., “content blocks”) so that only the \emph{style blocks} remain active for style-related embeddings.
    \item Training IP-Adapter from scratch: We followed the official training protocol on a 4M text-image paired dataset, but updated only the style blocks to prevent overfitting and excessive computation.
\end{enumerate}

Interestingly, both approaches delivered qualitatively similar stylized outputs---the full-scale training offered no substantial advantage over simply using the pre-trained style blocks. Hence, for convenience and computational efficiency, all subsequent experiments (Section~\ref{sec:exp}) are conducted with the \emph{pre-trained IP-Adapter} without further fine-tuning.

\noindent\textbf{IP-Adapter for Content \& Style.}
The ICAS framework that has been proposed adopts a two-branch strategy within the architecture of IP-Adapter: one branch handles content embedding injection, and the other handles style embedding injection. To ensure a stable and high-fidelity style transfer, we \emph{directly use the pre-trained IP-Adapter for style blocks} (i.e., \ the style injection path) without any further finetuning. This choice leverages large-scale training already performed in official IP-Adapter releases, guaranteeing robust style extraction and minimal computational overhead for new tasks.

\noindent\textbf{Partial Finetuning on Content Branch.}
While the style blocks remain frozen, we \emph{only update the content injection cross-attention sub-blocks} in IP-Adapter, focusing on a small dataset to adapt these content modules to multi-subject scenarios. Specifically:
\begin{itemize}
    \item Content Cross-Attention Sub-Block: 
    We train a few MLP layers and gating parameters in the content injection pathway to align the content embeddings with our target multi-subject distribution. The gating mechanism introduced in Section~\ref{sec:method} benefits from a modest amount of data to learn how to fuse content embeddings across diverse subject appearances.
    \item Frozen Style Blocks:
    The style injection path, which is responsible for projecting and injecting style embeddings from a reference image, is fully inherited from the official pre-trained IP-Adapter weights. The absence of large-scale supplementary training in this context contributes to the maintenance of stable style infusion.
\end{itemize}

\noindent\textbf{Motivation \& Outcome.}
By decoupling the style path (frozen) and the content path (lightly trained), we achieve two goals: (1) preserve the powerful generalization of the pre-trained style blocks, preventing style distortion that might arise from limited data, and (2) adapt the content cross-attention to multi-subject settings with minimal parameter updates. In practice, this hybrid strategy yields satisfactory stylization quality while ensuring robust multi-subject fidelity, as verified in Section~\ref{sec:exp}. It was also observed that full fine-tuning of the IP-Adapter or training of all blocks yielded similar or slightly less consistent results under limited data. This finding supports the \emph{effectiveness} and \emph{efficiency} of the partial fine-tuning approach.

\noindent\textbf{Base Model Setup.}
The Base Model Setup. ICAS was developed by employing a pre-trained text-to-image diffusion backbone, without the necessity of tuning its primary UNet parameters. A frozen CLIP text/image encoder was employed to generate text and style embeddings. Concurrently, a distinct structural encoder (e.g., for edges or depth) was integrated via ControlNet.

\noindent\textbf{Training Scheme.}
We only updated:
\begin{enumerate}
  \item The \emph{content injection branch} in the Style Injection Module (SIM), including gating parameters introduced in Section~\ref{sec:method}.
  \item The \emph{residual projection} in the Structure Preservation Module (SPM) for controlling structural alignment.
\end{enumerate}
All other parameters, including the IP-Adapter style path, were kept frozen to avoid overfitting and reduce computational overhead.

\noindent\textbf{Hyperparameters.}
The input images were resized to a fixed resolution (e.g.\ $512\times512$). We employed an AdamW optimizer with a constant learning rate for simplicity. A small $\lambda_{\text{gate}}$ was assigned to the gating regularizer to avoid extreme style vs.\ content weighting. The batch size and total iterations were set consistently across all datasets.

\noindent\textbf{Data Augmentation.}
To enhance the robustness of the multi-subject extraction and embedding processes, we occasionally applied random horizontal flips and mild color jitter to the content images. No substantial augmentation was performed on style references to maintain the integrity of style-specific characteristics.

\begin{figure}
    \centering
    \includegraphics[width=1\linewidth]{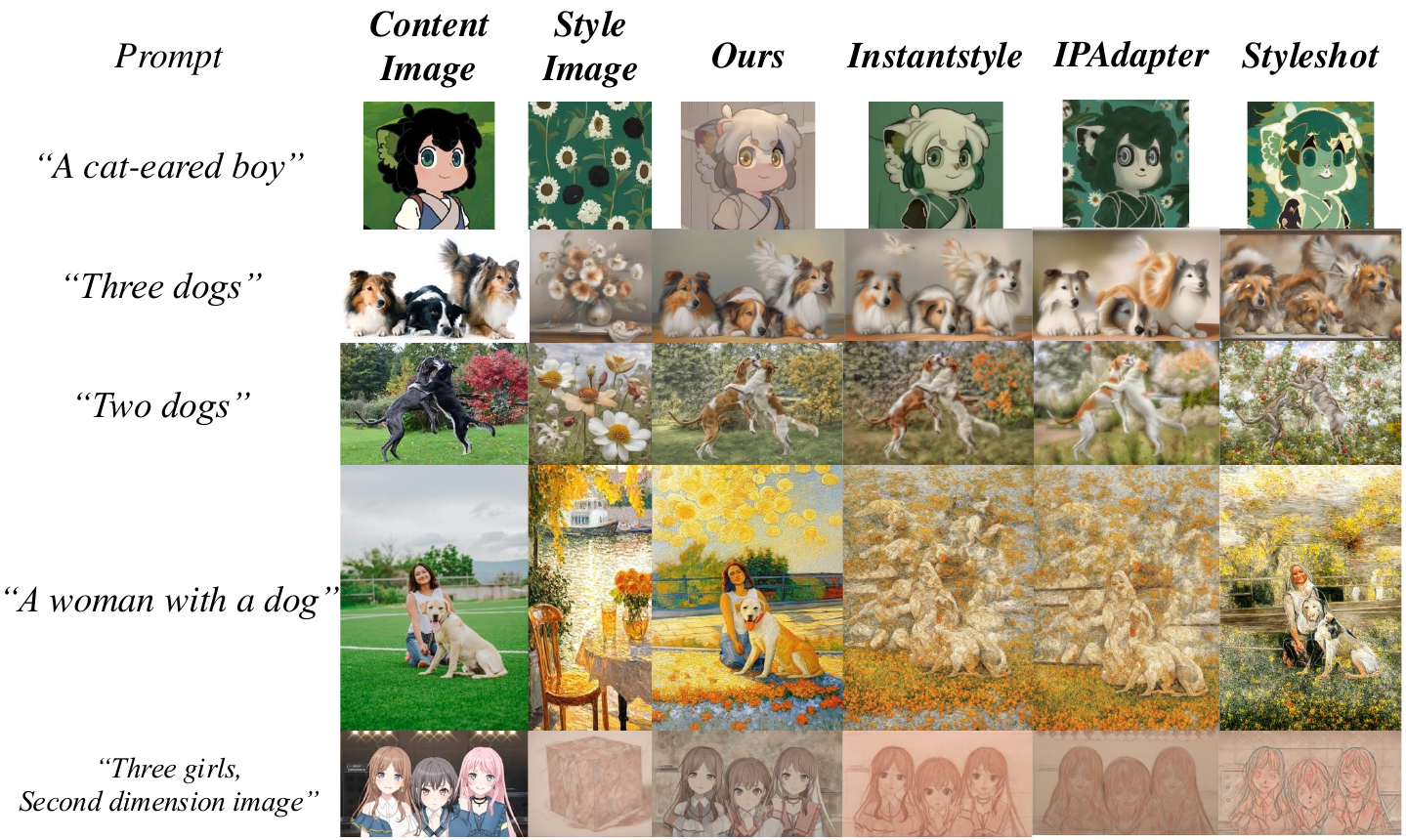}
    \caption{Comparison with state-of-the-art stylization approaches.}
    \label{fig:compareother}
\end{figure}

\subsection{Comparison with State-of-the-art Methods}
We evaluate our method against leading state-of-the-art stylization approaches, including InstantStyle, Styleshot, and the original IPAdapter with weight adjustment. In light of the intricacies involved in defining style and the variability in style definitions across models trained on their respective official datasets, a standardized comparison framework is implemented. To ensure fairness, the official default weights of each model are used, or the weights are manually adjusted to maintain consistent intensity sensitivity across all methods.

Figure~\ref{fig:compareother} shows the performance of each method on different types of sample images. Although all methods achieve some degree of style transfer, there are some key differences:
\begin{itemize}
\item \textbf{InstantStyle} uses a similar cue injection technique as IPAdapter, but can have difficulty distinguishing smaller subjects if the style texture is very detailed, and there is still some degree of recognition error for multiple subjects. While InstantStyle provides compelling stylization results in simpler scenes, they lack the flexibility to handle multiple subject viewpoints without artifacts.
\item \textbf{Styleshot} While very effective at reproducing the colors or textures of a painting, the official default weights have a greater impact on the content than the text.
\item \textbf{Original IPAdapter} performs relatively well on near-realistic inputs through weight adjustment, but may still be under-stylized or have some style "leakage" unless carefully tuned.
\end{itemize}
In contrast, the ICAS method preserves the relative layout of each subject while more consistently conveying the reference style. ICAS demonstrates robust performance even in complex backgrounds, a feat attributable to the fine-tuning of the criss-cross attention module within the content-graph pathway, complemented by multi-content embedding and ControlNet-based structural constraints.

A comprehensive evaluation of the results indicates that ICAS effectively achieves a balanced compromise between maintaining stylistic fidelity and preserving thematic integrity. The partial training design employed in this study involves the freezing of other modules while updating the criss-cross attention on the content adapter path. This approach has been shown to minimize computational overhead while achieving multi-disciplinary adaptability. ControlNet-based structural guidance has been demonstrated to prevent geometry collapse. Consequently, ICAS demonstrates superior performance in challenging content, exhibiting balanced color usage, accurate topic preservation, and stable geometry across diverse style references.

\begin{figure}
    \centering
    \includegraphics[width=1\linewidth]{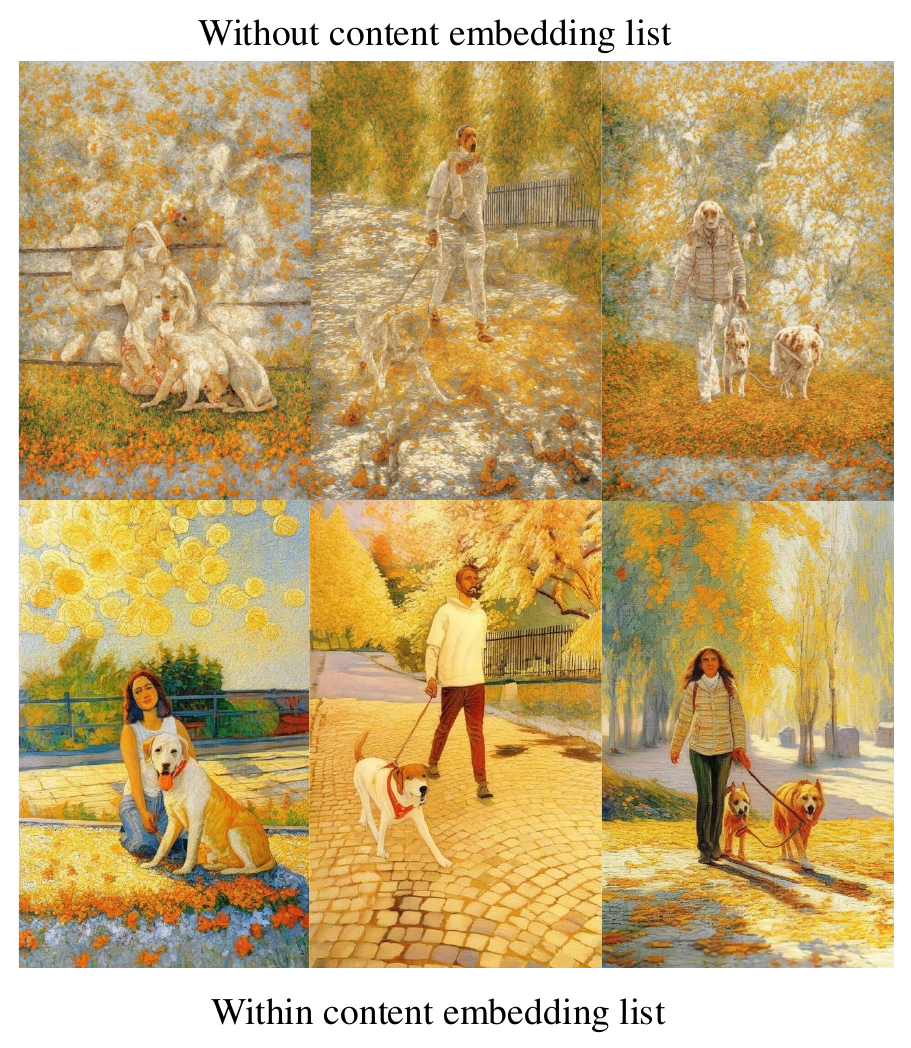}
    \caption{\textbf{The impact of multiple content embeddings on multi-subject fidelity. }
We compare \emph{single embedding} (encoding each content image only once) with \emph{multi-embedding} method (injecting multiple embeddings cyclically).
As shown in the figure, the single embedding strategy occasionally occludes or merges some subjects into the background, while the multi-embedding strategy can preserve different appearances for each subject.
This confirms that injecting multiple content embeddings can effectively improve the identity preservation rate of each subject in multi-subject scenarios.}
    \label{fig:ablate}
\end{figure}

\begin{figure}
    \centering
    \includegraphics[width=1\linewidth]{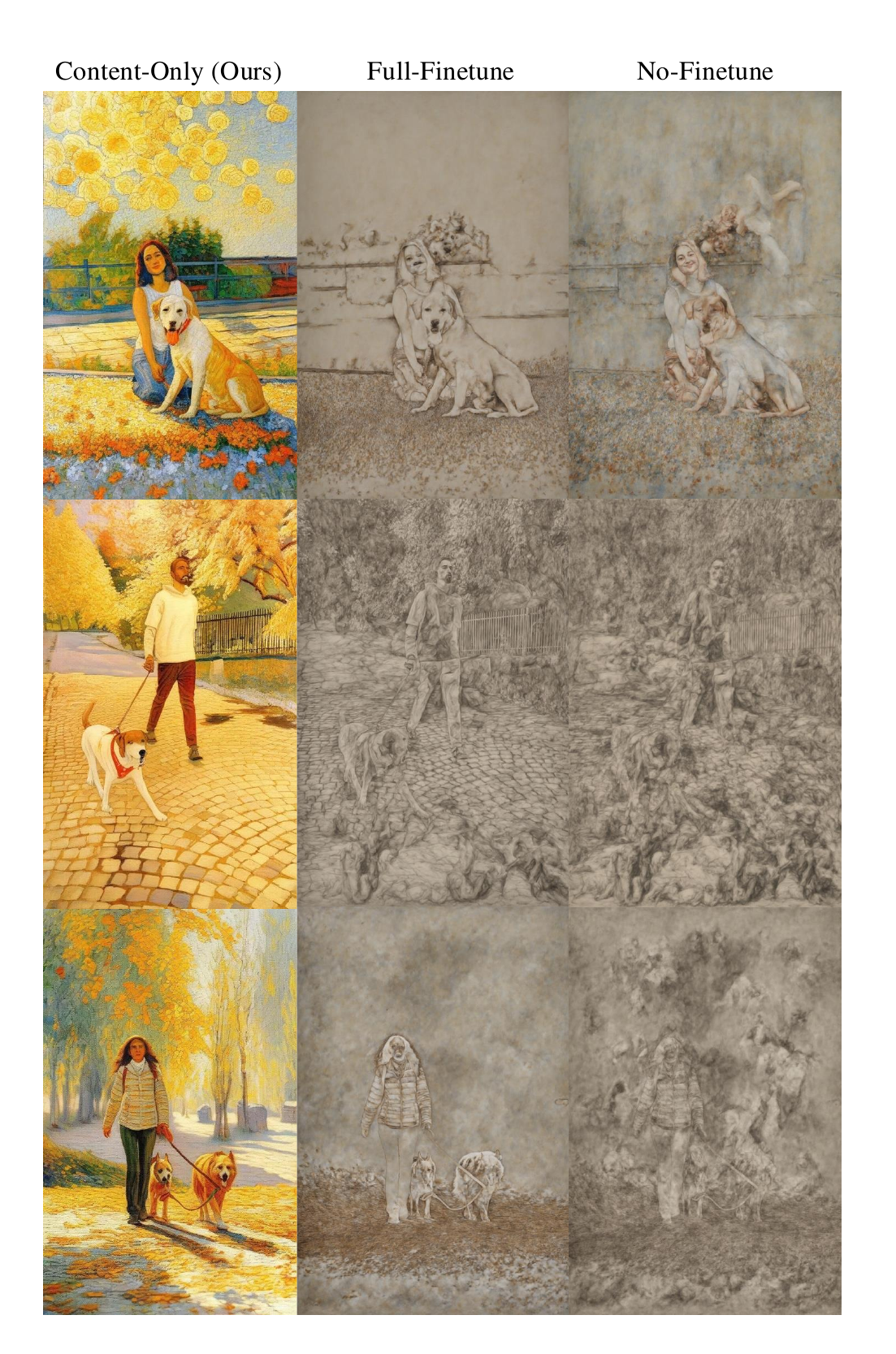}
    \caption{\textbf{Training strategies and complexity analysis}.
We compare three methods:
\emph{No-Finetune} (fully pre-trained IP adapter, no updating),
\emph{Full-Finetune} (simultaneously training style and content blocks),
and our \emph{Content-Only} method (freezing style blocks and only updating \texttt{num\_control\_attn}).
As shown in the figure, \emph{No-Finetune} and \emph{Full-Finetune} often suffer from style imbalance or inaccurate subject identification,
while the \emph{Content-Only} strategy retains the pre-trained style knowledge and can effectively adapt to multi-subject content.
This achieves the best balance between parameter overhead and final image fidelity.}
    \label{fig:tune}
\end{figure}

\begin{figure}
    \centering
    \includegraphics[width=1\linewidth]{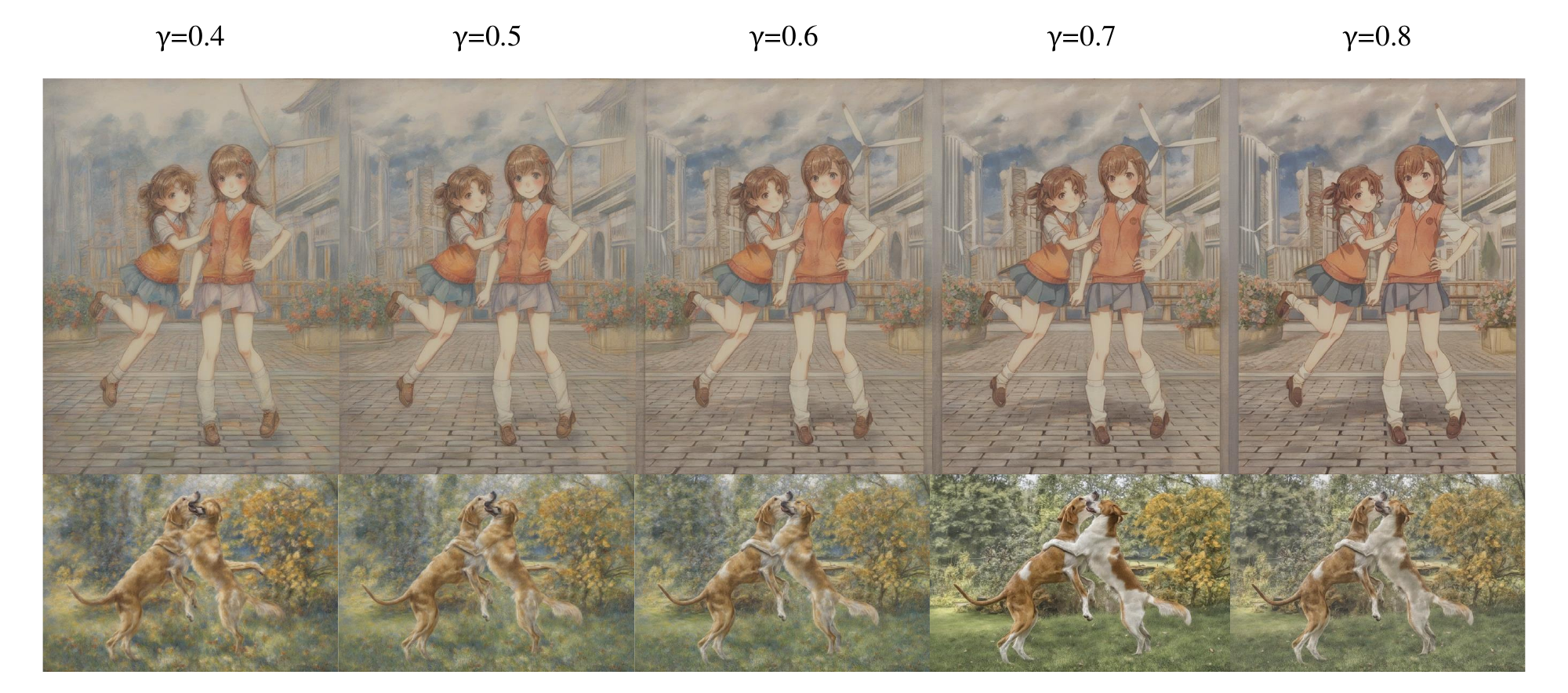}
    \caption{\textbf{Effect of structure scale \(\gamma\) on geometry and style.}
    From left to right, we vary \(\gamma\) from 0.4 to 0.8 when injecting ControlNet's structural features. 
    Lower values (e.g.\ 0.4 or 0.5) under-constrain geometry, causing slight subject distortions, 
    while values above 0.6 yield more stable layouts. 
    We empirically find \(\gamma = 0.7\) strikes the best balance between structure preservation and rich stylization.}
    \label{fig:structure_scale}
\end{figure}

\subsection{Ablation Studies and Analysis}  

Figure~\ref{fig:compareother} shows that the proposed ICAS method outperforms many state-of-the-art methods such as Instantstyle and Styleshot.
In what follows,  the proposed ICAS method is comprehensively analyzed from five aspects to investigate the logic behind its superiority.
(1) Role of the gating mechanism.
(2) Influence of multi-content embeddings.
(3) Training strategy and complexity.
(4) Effect of Structure Scale.

\subsubsection{Role of the Gating Mechanism}
We aim to verify whether the learnable or user-controllable scale factor (e.g.\ \(\alpha\) or \(\text{scale}\))—which acts as a gating term on style injection—truly balances style strength and content fidelity.
We compare:
\begin{itemize}
    \item \textbf{No-Gate}: replace the gating variable with a fixed constant (set to 1.0), effectively forcing style injection to be fully applied without adaptation.
    \item \textbf{Full ICAS (Ours)}: keep the scale/gating as defined in Eq.~(4) (or your code’s \(\text{scale}\) usage), allowing it to adjust style intensity dynamically.
\end{itemize}

The present study evaluates the effects of each variant on structure preservation and style accuracy in multi-subject images. The experimental results indicate that the elimination of the gating mechanism results in the oversimplification of complex scenes, frequently leading to the obscuring of secondary subjects. In contrast, the version with gating more effectively preserves subject boundaries while injecting a recognizable style. This finding indicates that the gating mechanism, even if solely represented as a "scale" parameter in the code, plays a pivotal role in balancing style injection. Fixing it to 1 or 0 is inadequate for adapting to the diverse content scenarios that arise.

\subsubsection{Influence of Multi-Content Embedding}
We examine whether employing multiple content embeddings and cyclically injecting them truly benefits multi-subject fidelity.
\begin{itemize}
    \item \textbf{Single-Embed}: each content image is encoded only once, ignoring potential subject variety.
    \item \textbf{Multi-Embed (Ours)}: for the same content image, multiple embeddings are extracted (via augmentation or segmentation) and injected in a loop through self.num\_control\_attn.
\end{itemize}

We use a subject-level matching score, verifying whether each subject’s appearance is preserved after stylization. Figure~\ref{fig:ablate} shows that Single-Embed occasionally hides part of the subject in the background. Therefore, in a multi-subject scenario, injecting multiple embeddings helps the model distinguish each topic.

\subsubsection{Training Strategy and Complexity}
We investigate the effect of partially training only content-injection cross-attention versus fully training all IP-Adapter blocks, plus the implications for the overall complexity.
We compare:
\begin{itemize}
    \item \textbf{Content-Only (Ours)}: freeze style blocks, updating only num\_control\_attn parameters,
    \item \textbf{Full-Finetune}: train both style and content sub-blocks,
    \item \textbf{No-Finetune}: use only pre-trained IP-Adapter, no parameter updates.
\end{itemize}
We measure parameter count, training time, and final quality (FID / style match / subject fidelity). As shown in the figure~\ref{fig:tune}, both Full-Finetune and No-Finetune show style imbalance and inaccurate subject recognition. We also found repeated diffusion in some data. Currently, micro-training crossattention on the IPAd pater route of the content graph is the best approach: it preserves the pre-trained style knowledge while adapting to multi-subject content with minimal cost.

\subsubsection{Effect of Structure Scale}
We examine how the ControlNet conditioning scale, denoted by $\gamma$, influences the retention of multi-subject geometry and overall style quality. By varying $\gamma$ from 0.4 to 0.8, we aim to find a balanced sweet spot where the structure guidance is strong enough to maintain positional fidelity without overly dampening the stylization.

We fix all other parameters in ICAS (including the gating factor, multi-content embeddings, and partial training strategy). We then evaluate five values of $\gamma \in \{0.4, 0.5, 0.6, 0.7, 0.8\}$ on a subset of multi-subject test images. Figure~\ref{fig:structure_scale} displays representative results.
\begin{itemize}
    \item \textbf{\boldmath$\gamma=0.4$ or $0.5$}: The structural guidance is somewhat underpowered, leading to minor geometric drift or overlapping subjects in cluttered scenes.
    \item \textbf{\boldmath$\gamma=0.6$ -- $0.8$}: Structure is preserved more robustly, with fewer artifacts and more stable subject boundaries.
    \item \textbf{\boldmath\(\gamma=0.7\)}: Empirically yields the best trade-off between geometry and style fidelity, preventing over-constraint (which can dull style expression) while avoiding subject distortions.
\end{itemize}

Based on these experiments, we adopt $\gamma=0.7$ as our default structure scale in subsequent experiments, as it consistently balances multi-subject clarity with coherent style transfer. This observation underscores the importance of tuning ControlNet’s strength to neither overshadow the style injection nor under-enforce the spatial layout.

The findings from the comprehensive ablation experiments demonstrate the efficacy of each essential component in the proposed approach. The gating mechanism is critical to maintaining a balance between style injection strength and content fidelity. Its removal can lead to oversaturation and loss of detail in complex scenes. The multi-content embedding strategy has been shown to significantly improve subject preservation in multi-subject scenes by enabling the model to better distinguish individual elements. Finally, our targeted training approach—focusing only on the content-injected cross-attention mechanism while freezing other modules—achieves the best balance between preserving pre-trained style knowledge and adapting to multi-subject content with minimal computational overhead. These findings indicate that the design choices employed effectively address the challenge of maintaining style accuracy and content integrity during image stylization.

\begin{table}[t]
    \centering
    \caption{\textbf{User study results on multi-subject style transfer.}
    We report average scores (1–5) on three criteria: Style Fidelity, Subject Clarity, and Overall Aesthetic.
    \textbf{ICAS} consistently outperforms the baselines, indicating stronger style preservation and clearer multi-subject rendering in the eyes of participants.}
    \begin{tabular}{cccc}
        \toprule[1.5pt]
         &  Style Fidelity&  Subject Clarity&  Overall Aesthetic\\
         \midrule 
         IPAdapter&  3.92&  3.66&  3.79\\
         InstantStyle&  4.22&  3.84&  4.03\\
         ICAS (Ours)&  4.32&  4.28&  4.30\\
         \bottomrule 
         
    \end{tabular}
    \label{tab:user_study}
\end{table}

\subsection{User Study}\label{sec:user_study}
\noindent\textbf{Setup.}
Since style is hard to define, we decided to regress the evaluation of our style images to the “people” themselves. To do this, we conducted a small-scale user study. We randomly sampled 50 multi-subject content images and paired each with 3 distinct style references (yielding 150 total stylization tasks). We then generated results using:
\begin{itemize}
    \item \textbf{Baseline A}: Original IPAdapter
    \item \textbf{Baseline B}: InstantStyle
    \item \textbf{ICAS (Ours)}
\end{itemize}
All images were shown side-by-side in random order to participants without disclosing which method produced which image.

\noindent\textbf{Evaluation Criteria.}
Participants (20 people, each with some knowledge of AIGC image generation and some aesthetic sense) were asked to rate each image from 1 (worst) to 5 (best) according to the following aspects:
\begin{itemize}
    \item \textbf{Style Fidelity}: how well the output reflects the target style’s color palette, texture, and artistic traits.
    \item \textbf{Subject Clarity}: how distinct and recognizable each main subject remains, especially smaller or overlapping subjects.
    \item \textbf{Overall Aesthetic}: the overall impression of the stylized result, balancing style correctness with visual coherence.
\end{itemize}

\noindent\textbf{Results.}
Table~\ref{tab:user_study} reports the average scores. Our \textbf{ICAS} approach consistently ranks higher in both \emph{Style Fidelity} (mean 4.32) and \emph{Subject Clarity} (mean 4.28) compared to Baseline A (3.92/3.66) and Baseline B (4.22/3.84). The participants noted that ICAS “maintains a better sense of the shape and limits of each subject while still looking stylistically pleasing.” In \emph{Overall Aesthetic}, ICAS also achieved a mean score of 4.30, overshadowing the baseline methods (3.79 and 4.03 respectively).

\noindent\textbf{Discussion}
These subjective ratings corroborate our quantitative findings: the integration of IPAdapter and ControlNet, multi-content embedding, and partial training yields visually compelling and well-defined results in multi-topic style transfer.

\begin{figure}
    \centering
    \includegraphics[width=1\linewidth]{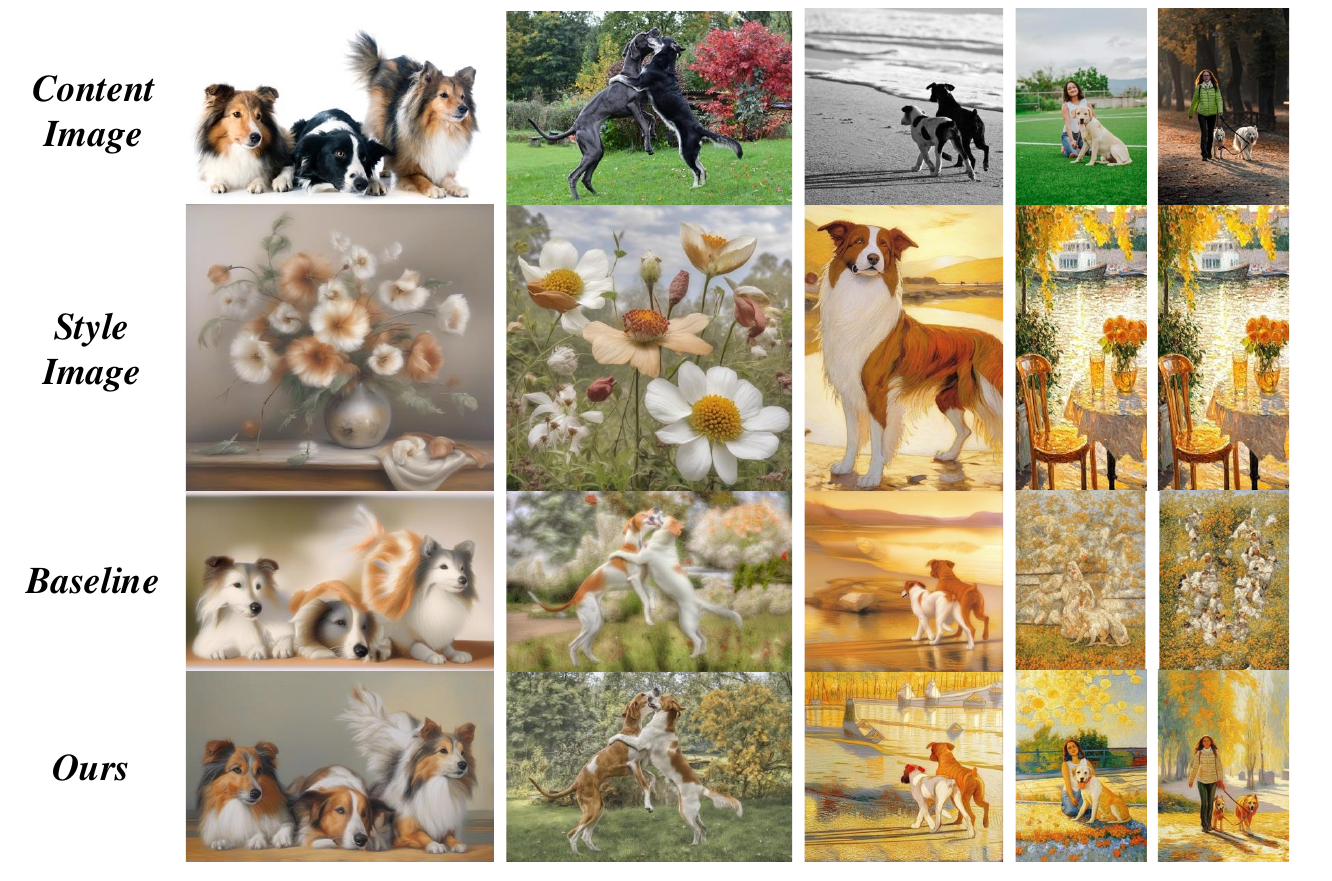}
    \caption{\textbf{Qualitative comparisons for multi-subject style transfer.}
    Each example shows four rows: 
    (1)~the ground-truth content or reference image, 
    (2)~the style image, 
    (3)~output from a strong baseline approach, 
    and 
    (4)~our ICAS result. 
    While the baseline often struggles to apply style details consistently or loses smaller subjects (especially in cluttered backgrounds), our ICAS framework effectively retains each subject’s identity and achieves high-fidelity style transfer. This improvement stems from leveraging multiple content embeddings (via the IPAdapter pathway) and ControlNet-based structure injection, along with a partial fine-tuning strategy and gating mechanism that together balance color palette fidelity and multi-subject geometry.}
    \label{fig:vis_compare}
\end{figure}

\subsection{Visualization}  
\noindent

Beyond quantitative metrics, we provide visual comparisons to qualitatively assess the performance of different approaches in multi-subject style transfer. Figure~\ref{fig:vis_compare} illustrates three rows of images for each example:

\begin{itemize}
    \item Row 1 (GT): The ground-truth content or reference image for context.
    \item Row 2 (Style): Provide corresponding style information.
    \item Row 3 (Baseline): Outputs from a representative baseline method (\textit{e.g.}, a tuning-free adapter or a fine-tuning approach). We select the best-performing baseline among the state-of-the-art for clarity.
    \item Row 4 (Ours): The stylized result generated by our proposed ICAS framework, preserving multi-subject semantics while transferring the reference style.
\end{itemize}

From the figure, we can observe the following:
\begin{enumerate}
    \item \textbf{Style Fidelity}: While the baseline sometimes exhibits diluted or incomplete style attributes (especially in more complex scenes), our result more consistently reflects the target color palette and textures.
    \item \textbf{Multi-Subject Preservation}: The Baseline approach tends to mix or lose smaller subjects in crowded scenes. By contrast, our approach maintains each subject’s identity and boundaries, demonstrating effectiveness of multi-content embedding and ControlNet-based structure injection.
    \item \textbf{Cluttered Backgrounds}: In challenging backgrounds with overlapping objects, Baseline may create noticeable artifacts or misalignments, whereas ICAS retains more stable geometry.
\end{enumerate}

These visual comparisons serve to reinforce the key innovations inherent to ICAS. Specifically, the combination of multiple content embeddings from the iPAdapter pathway with ControlNet-based structural constraints has enabled the preservation of subtle distinctions between multiple subjects, while ensuring the maintenance of coherent, high-fidelity style transfer. Typically, baseline models demonstrate deficiencies in consistently applying style details or properly differentiating smaller subjects, particularly against complex backgrounds. ICAS, however, exhibits a balanced approach in its effectiveness in maintaining color fidelity and geometric stability. The findings underscore the synergistic impact of our partial training strategy, which incorporates the freezing of style blocks while concurrently updating content cross-attention mechanisms, in conjunction with our gating mechanism. This integrated approach ensures precise control over both style and multi-subject layout.

\section{Conclusion}\label{sec:con} 
We present ICAS, a novel framework for multi-subject style transfer that combines the strengths of IP-Adapter for flexible style injection and ControlNet for structural guidance. By selectively fine-tuning only the content injection branch of a pre-trained diffusion model, ICAS achieves a compelling balance between style fidelity and multi-subject semantic preservation. The proposed cyclic content embedding strategy further enables consistent appearance transfer without relying on large-scale stylized datasets.
Extensive experiments demonstrate that ICAS outperforms existing approaches in structural preservation, style coherence, and inference efficiency. However, limitations remain in content embedding alignment under highly complex scenes, and the recursive injection strategy incurs additional inference cost. In future work, we plan to explore lightweight content encoding schemes and extend ICAS to support more diverse subject categories and stylistic variations, promoting broader applicability in real-world generation tasks.

\bibliographystyle{IEEEtran}
\bibliography{main}

\newpage

\EOD

\end{document}